\documentclass[10pt,twocolumn,letterpaper]{article}

\usepackage{iccv}
\usepackage{times}
\usepackage{epsfig}
\usepackage{graphicx}
\usepackage{amsmath}
\usepackage{amssymb}

\usepackage{color}
\usepackage{cite}
\usepackage{array}
\usepackage{tabularx}
\usepackage{setspace}
\usepackage{url}
\usepackage{booktabs}
\usepackage{algorithm}
\usepackage{algorithmic}

\iccvfinalcopy 

\def\iccvPaperID{****} 

\begin{document}

\title{How to Train Triplet Networks with 100K Identities?}

\author{Chong Wang\\
Orion Star\\
Beijing, China\\
{\tt\small chongwang.nlpr@gmail.com}
\and
Xue Zhang\\
Orion Star\\
Beijing, China\\
{\tt\small yuannixue@126.com}
\and
Xipeng Lan\\
Orion Star\\
Beijing, China\\
{\tt\small xipeng.lan@gmail.com}
}

\maketitle

\begin{abstract}
    Training triplet networks with large-scale data is challenging in face recognition.
    Due to the number of possible triplets explodes with the number of samples, previous studies adopt the online hard negative mining(OHNM) to handle it.
    However, as the number of identities becomes extremely large, the training will suffer from bad local minima because effective hard triplets are difficult to be found.
    To solve the problem, in this paper, we propose training triplet networks with subspace learning, which splits the space of all identities into subspaces consisting of only similar identities.
    Combined with the batch OHNM, hard triplets can be found much easier.
    Experiments on the large-scale MS-Celeb-1M challenge with $100K$ identities demonstrate that the proposed method can largely improve the performance.
    In addition, to deal with heavy noise and large-scale retrieval, we also make some efforts on robust noise removing and efficient image retrieval, which are used jointly with the subspace learning to obtain the state-of-the-art performance on the MS-Celeb-1M competition (without external data in $Challenge1$).
\end{abstract}

\section{Introduction}
    Triplet loss is a metric learning method that has been widely used in many applications, \emph{e.g.}, face recognition\cite{cnn_triplet_facenet,cnn_vggnet}, image retrieval\cite{cnn_triplet_lift_structure,cnn_deep_retrieval} and person re-identification\cite{cnn_triplet_defense,cnn_triplet_multi-task}.
    A triplet usually consists of three samples: an anchor sample, a positive one with the same class to the anchor, and a negative one with the different class.
    The objective of triplet loss is to learn a metric that pushes the positive pairs closer while pulls the negative pairs away, thus the samples within the same class can be nearest to each other.
\begin{figure}[!htb]
\centering
\includegraphics[width=3in]{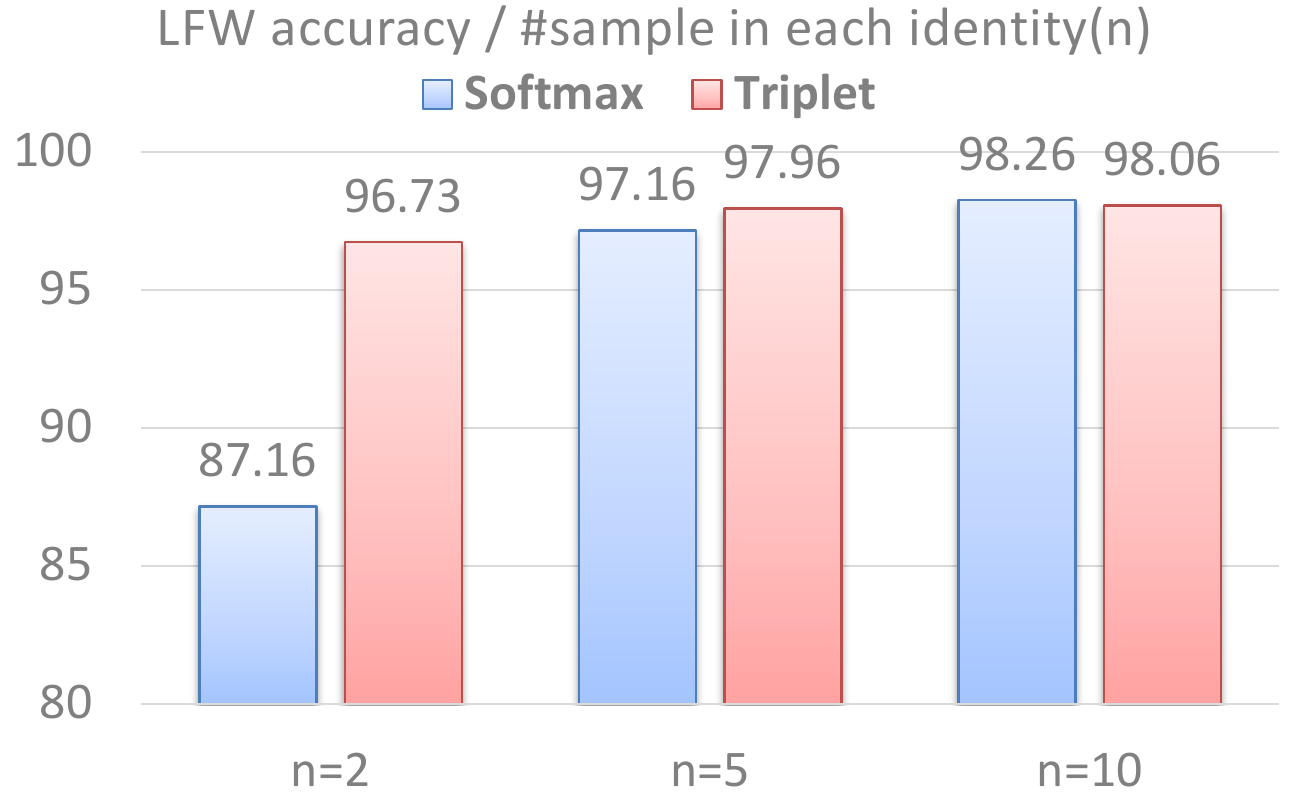}
\caption{The LFW accuracy of the models trained with softmax and triplet loss with the number of samples in each class set to be 2, 5 and 10. The model is trained on MS-Celeb-1M\cite{dataset_msraface} with $100K$ identities, and the evaluation is given by LFW\cite{dataset_lfw}.}
\label{fg:why_triplet}
\vspace{-0.4cm}
\end{figure}

    One question is why we need the triplet loss?
    Actually, there are some alternatives such as the softmax loss, which is also popular.
    However, as the number of classes becomes extremely large, the fully-connected layer that connects to softmax loss will also become extremely large, thus the GPU memory cost will be unbearable with an usual batchsize, while the small batchsize will take too long to train.
    Another reason is that if only a few samples are available in each class, training with softmax loss is difficult, and Fig.\ref{fg:why_triplet} shows its influence on softmax and triplet loss with $100K$ identities.
    The triplet performs much better when the number of samples in each class is small($n=2$).

    Though the advantage is clear, there are some challenges to use it.
    One big challenge is how to train triplet models effectively with large-scale data, \emph{e.g.}, $100K$ and $1M$ identities are common cases in face recognition.
    The difficulty of scaling triplet is that the number of possible triplets is cubic in the number of samples, and most triplets are too easy that cannot help training.
    To reduce searching space, some researchers transferred the triplet loss into softmax loss\cite{cnn_triplet_to_binary_cls,cnn_triplet_to_match_cls,cnn_triplet_to_probilistic_embed_cls}, while some proposed the Online Hard Negative Mining(OHNM)\cite{cnn_triplet_early_online,cnn_triplet_vggface,cnn_triplet_facenet} or batch OHNM\cite{cnn_triplet_defense}.
    Most studies focused on the small-scale case, while FaceNet\cite{cnn_triplet_facenet} used an extremely large number of identities($8M$), but it suffered from long training time(a few months).

    In the above methods, all of them consider all identities to sample the batch.
    It is widely accepted that hard triplets can help training because they can reduce the ambiguity of recognizing similar identities, and it indicates these hard triplets should better come from similar identities.
    However, sampling from all the identities cannot guarantee to include the similar ones, thus it will fail in generating effective hard triplets.
    Especially, in the large-scale face recognition with $100K$ or $1M$ identities, the probability of sampling similar identities with an usual batchsize is very tiny, \emph{e.g.}, the batch with the size of $1800$ is randomly sampled from $8M$ identities in FaceNet\cite{cnn_triplet_facenet}.
    Therefore, how to find the similar identities is the key to improve the training of triplet networks with large-scale data.

    In this paper, we consider the case of large-scale face recognition, and propose to train triplet networks with subspace learning.
    The basic idea is to generate a representation for each identity, and apply clustering on all the identities to generate clusters or subspaces, wherein identities are similar in each subspace.
    With the batch OHNM applied in each subspace iteratively, the proposed method can easily generate more effective hard triplets.
    Evaluations on the large-scale MS-Celeb-1M dataset with $100K$ identities\cite{dataset_msraface} show that the proposed method can largely improve performance and get more robust triplet models.
    Particularly, our single triplet network obtains the LFW\cite{dataset_lfw} accuracy of $99.48\%$, which can be competitive to FaceNet's $99.63\%$\cite{cnn_triplet_facenet} with $8M$ identities.

    This subspace learning with batch OHNM is our main contribution.
    In addition, given the fact that the MS-Celeb-1M dataset is noisy, we also design a noise removing trick to clean the training data at the beginning, and experiments show that it is able to handle different number of noises.
    Furthermore, consider that the number of training images is about $5M$ after the cleaning, we use a two-layer hierarchical trick to retrieve a test image accurately and efficiently.
    Combined with the proposed triplet network, we have achieved the state-of-the-art performance on the MS-Celeb-1M competition, \emph{i.e.}, $Challenge1$ without external data.
    The recall of the random and hard set is $75\%$ and $60.6\%$ respectively, in which $75\%$ has achieved the upper limit on the random set.

\section{Related Work}
    In this part, we introduce some previous studies on how to accelerate the training of triplet networks.
    One big difficulty is that the number of possible triplets scales cubically with the number of training samples.
    To avoid directly searching the whole space, some researchers\cite{cnn_triplet_to_binary_cls,cnn_triplet_to_match_cls,cnn_triplet_to_probilistic_embed_cls} convert the triplet loss into a form of softmax loss.
    Sankaranarayanan \etal.\cite{cnn_triplet_to_probilistic_embed_cls} propose to transfer the Euclidean distance between positive and negative pairs into probabilistic similarity, and they use low-dimensional embedding for fast retrieval.
    Similar to \cite{cnn_triplet_to_probilistic_embed_cls}, Zhuang \etal.\cite{cnn_triplet_to_binary_cls} convert the retrieval problem into a multi-label classification problem with binary codes, which is optimized by the binary quadratic algorithm to achieve faster retrieval.
    To simplify the optimization, Hoffer \etal.\cite{cnn_triplet_to_match_cls} propose a Siamese-like triplet network by transferring the retrieval problem into a $2$-class classification problem.
    These methods have shown promising speedup, but no hard triplets are considered, which will result in inferior performance.

    Inspired by the efficiency of classification, some studies\cite{cnn_triplet_early_online,cnn_triplet_vggface,cnn_triplet_multi-task,cnn_triplet_facenet} combine the advantages of classification and hard triplets.
    Wang \etal.\cite{cnn_triplet_early_online} use a pretrained classification model to select possible hard triplets offline, but the offline selection is fixed as the classification model will not be updated.
    To achieve faster training and handle variant triplets, Parkhi \etal.\cite{cnn_triplet_vggface} train a classification network that is further fine-tuned with triplet loss.
    They use the Online Hard Negative Mining(OHNM), wherein only the triplets violating the margin constraint are considered as the hard ones for learning.
    Instead of fine-tuning with only triplet loss, Chen \etal.\cite{cnn_triplet_multi-task} propose to train networks jointly with softmax and triplet loss to preserve both inter-class and intra-class information, and they also adopt OHNM in training.
    To apply OHNM in large-scale data, Schroff \etal. propose the FaceNet\cite{cnn_triplet_facenet} that trains triplet networks with $8M$ identities, and it takes a few months to finish with a large batchsize of $1800$.
    One limitation of OHNM is that triplets are predefined in the batch, and this will miss possible hard negative samples that contained in the batch.

    To make full use of the batch, some studies\cite{cnn_triplet_defense} generate hard triplets online within the batch.
    Hermans \etal.\cite{cnn_triplet_defense} propose the batch OHNM, in which negative samples are searched within the batch based on their distance to the anchor, and the top nearest ones are considered as candidate hard negative samples.
    In this way, more hard triplets can be found easily, and the best performance is obtained in person re-identification with $1500$ identities.
    Due to their small scale, the probability of sampling similar identities with an usual batchsize($128$ or $256$) is large.
    However, in the large-scale case, randomly sampling similar identities will be much more difficult, thus the batch OHNM will fail.
    In this paper, we focus on how to find effective hard triplets in large-scale face recognition.

\section{Method}
    In this part, we will introduce how to train triplet networks with large-scale data.
    We first review the Online Hard Negative Mining(OHNM) and batch OHNM in training the triplet network, then we elaborate how to improve the training with subspace learning.
\begin{figure*}[!htb]
\centering
\includegraphics[width=6in]{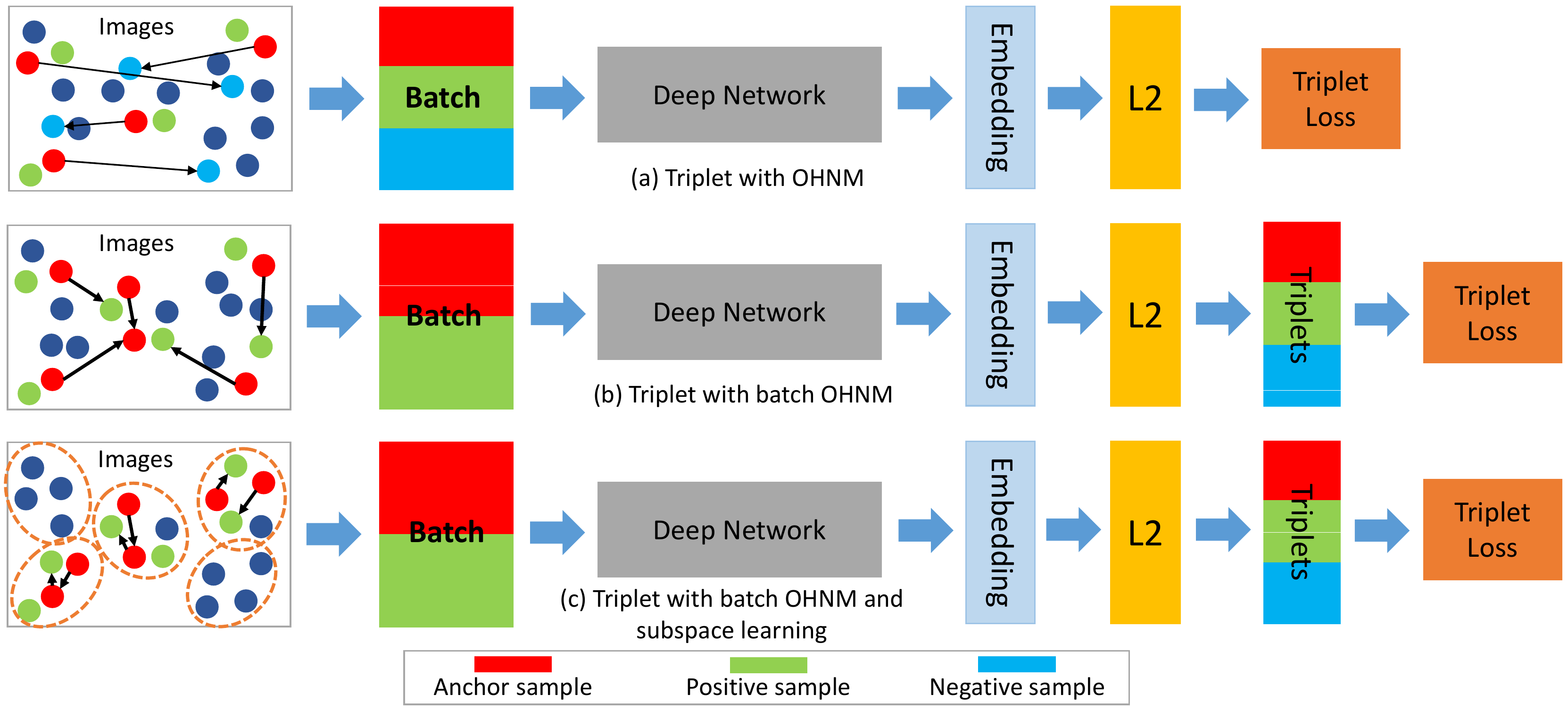}
\caption{The pipeline of training triplet networks with three methods. (a) Triplet with OHNM; (b) Triplet with batch OHNM; (c) Triplet with batch OHNM and subspace learning.}
\label{fg:triplet_framework}
\vspace{-0.3cm}
\end{figure*}

    Let $x$ be an image.
    Similar to FaceNet\cite{cnn_triplet_facenet}, we map the sample $x$ to a $d$-dimensional embedding with $L2$-normalization, as shown in Fig.\ref{fg:triplet_framework}(a), and this gives the representation $f\left( x \right) \in {\mathbb{R}^d}$ that satisfies ${\left\| {f\left( x \right)} \right\|^2} = 1$.
    Let ${x_i^a}$, ${x_i^p}$ and ${x_i^n}$ be the anchor sample, positive sample and negative sample respectively, in which ${x_i^a}$ and ${x_i^p}$ have the same identity, while ${x_i^a}$ and ${x_i^n}$ come from different identities, \emph{i.e.}, ${\rm I}\left( {x_i^a} \right) = {\rm I}\left( {x_i^p} \right)$ and ${\rm I}\left( {x_i^a} \right) \ne {\rm I}\left( {x_i^n} \right)$, wherein $I\left( x \right)$ denotes the identity of $x$.

\subsection{Triplet with OHNM}
    Online Hard Negative Mining(OHNM)\cite{cnn_triplet_early_online,cnn_triplet_vggface,cnn_triplet_multi-task,cnn_triplet_facenet} is proposed to only focus on hard triplets for training, and the triplet loss with OHNM can be formulated as minimizing the following loss
\begin{equation}
\label{eqn_triple_ohem}
\renewcommand\arraystretch{1}
  \begin{gathered}
  {\sum\nolimits_{i = 1}^{\left| B \right|} {\left[ {{{\left\| {f\left( {x_i^a} \right) - f\left( {x_i^p} \right)} \right\|}^2} - {{\left\| {f\left( {x_i^a} \right) - f\left( {x_i^n} \right)} \right\|}^2} + \alpha } \right]} _ + } \hfill \\
  \forall x_i^a,x_i^p,x_i^n \in T \hfill \\
  \end{gathered},
\end{equation}
    wherein $\left| B \right|$ is the batchsize, $B$ is the batch with $3\left| B \right|$ images sampled from the image space $T$, and $\alpha$ denotes the margin enforced between positive and negative pairs.
    In Eqn.\ref{eqn_triple_ohem}, hard triplets are the ones that violate the margin constraint.
    However, due to ${x_i^n}$ is randomly sampled from all identities, it is difficult to generate effective hard triplets.
    Fig.\ref{fg:triplet_framework}(a) gives an illustration, where most sampled ${x_i^n}$(light blue circles) are far away from the positive pairs.
    Though the hard negative ones are in the batch, they cannot be selected, \emph{e.g.}, some dark blue circles.

\subsection{Triplet with Batch OHNM}
    To fully exploit the batch, some researchers propose the batch OHNM\cite{cnn_triplet_defense} to explore more negative samples.
    Instead of sampling ${x_i^n}$ from the whole image space $T$, batch OHNM finds ${x_i^n}$ in the batch, and the loss with batch OHNM is minimized as follows
\begin{equation}
\label{eqn_triple_batch_ohem}
\renewcommand\arraystretch{1}
  \begin{gathered}
  {\sum\nolimits_{i = 1}^{\left| B \right|} {\left[ {{{\left\| {f\left( {x_i^a} \right) - f\left( {x_i^p} \right)} \right\|}^2} - {{\left\| {f\left( {x_i^a} \right) - f\left( {x_i^n} \right)} \right\|}^2} + \alpha } \right]} _ + } \hfill \\
  s.t.\;x_i^n = \mathop {\arg \min }\limits_{{x}} {\left\| {f\left( {x_i^a} \right) - f\left( {{x}} \right)} \right\|^2},\;{\rm I}\left( {x_i^a} \right) \ne {\rm I}\left( {{x}} \right) \hfill \\
  \forall x_i^a,x_i^p \in T,\;{x} \in B \hfill \\
  \end{gathered}.
\end{equation}
    Different from Eqn.\ref{eqn_triple_ohem}, ${x_i^n}$ is selected with respect to the Euclidean distance to each anchor ${x_i^a}$.
    In training, we randomly sample $\left| B \right|$ different identities, wherein ${x_i^a}$ and ${x_i^p}$ are also randomly sampled in each identity, thus there are $2\left| B \right|-2$ possible ${x_i^n}$ for each ${x_i^a}$.
    Batch OHNM is more advantageous not only for the harder negatives, but also more triplets can be used for training as ${x_i^n}$ is no longer the input.
    Similar to FaceNet\cite{cnn_triplet_facenet}, we consider several nearest ${x_i^n}$ but not the most nearest one, because it might lead to poor training as mislabelled and poorly imaged faces would dominate ${x_i^n}$.
    Fig.\ref{fg:triplet_framework}(b) gives an illustration, in which some ${x_i^n}$ are much closer to positive pairs, and more hard triplets can be used to accelerate training, \emph{e.g.}, $6$ triplets in Fig.\ref{fg:triplet_framework}(b) compared to $4$ triplets in Fig.\ref{fg:triplet_framework}(a).

\subsection{Triplet with Subspace Learning}
    The effectiveness of hard negative mining comes from its ability to handle ambiguity in recognizing similar identities, and this indicates the hard triplets should better be generated from similar identities.
    However, in OHNM and batch OHNM, all identities are used to randomly sample the batch, which cannot be guaranteed to include similar identities.
    Especially in the large-scale case with $100K$ identities, sampling similar identities with an usual batchsize such as $128$ or $256$ can be rather difficult.

    To find similar identities, the basic idea is to get identity representation and cluster on all identities to generate subspaces, wherein identities can be similar.
    We achieve this with a classification model, which can be pretrained on a subset of the whole training set.
    Let $x_i^c\left( {\forall i = 1,...,{N_c}} \right)$ be an image with the identity $c$, and ${{N_c}}$ is the number of images in that identity.
    Then, the representation of $x_i^c$ extracted by the classification model is denoted as $g\left( {x_i^c} \right)$, and the identity representation ${g\left( c \right)}$ is given by
\vspace{-0.1cm}
\begin{equation}
\label{eqn_identity_representation_cls}
\renewcommand\arraystretch{1}
   {{g\left( c \right) = \sum\nolimits_{i = 1}^{{N_c}} {g\left( {x_i^c} \right)} } \mathord{\left/
   {\vphantom {{g\left( c \right) = \sum\nolimits_{i = 1}^{{N_c}} {g\left( {x_i^c} \right)} } {{N_c}}}} \right.
   \kern-\nulldelimiterspace} {{N_c}}},\;\forall c = 1,...,C.
\end{equation}
    Then, we apply k-means clustering on all the identity representation $\left[ {g\left( 1 \right),...,g\left( C \right)} \right]$ to generate $M$ subspaces, and each subspace is denoted as ${T_m}\left( {\forall m = 1,...,M} \right)$, as the dotted circles shown in Fig\ref{fg:triplet_framework}(c).

    To accelerate training, we refer to \cite{cnn_triplet_vggface} that uses the pretrained classification model as initialization.
    With batch OHNM applied in each subspace iteratively, the triplet loss with subspace learning can be minimized as
\begin{equation}
\label{eqn_triple_batch_ohem_subspace}
\renewcommand\arraystretch{1}
  \begin{gathered}
  {\sum\nolimits_{i = 1}^{\left| B \right|} {\left[ {{{\left\| {f\left( {x_i^a} \right) - f\left( {x_i^p} \right)} \right\|}^2} - {{\left\| {f\left( {x_i^a} \right) - f\left( {x_i^n} \right)} \right\|}^2} + \alpha } \right]} _ + } \hfill \\
  s.t.\;x_i^n = \mathop {\arg \min }\limits_{{x}} {\left\| {f\left( {x_i^a} \right) - f\left( {{x}} \right)} \right\|^2},\;{\rm I}\left( {x_i^a} \right) \ne {\rm I}\left( {{x}} \right) \hfill \\
  \forall {x} \in B,\;x_i^a,x_i^p \in {T_m},\;m = 1,...,M \hfill \\
  \end{gathered}.
\end{equation}
    Different from the single batch OHNM, the batch is randomly sampled in ${T_m}$ with similar identities, thus the selected ${x_i^n}$ can be much harder to generate more effective hard triplets.
    Fig.\ref{fg:triplet_framework}(c) illustrates this process, wherein hard negative samples are very close to the positive pairs in a subspace(dotted circle).
    Though some time is cost in feature extraction and clustering, the subspace learning can largely reduce the searching space and training time.
    Particularly, similar to batch OHNM, several nearest ${x_i^n}$ are considered to avoid bad local minima.

\subsection{Some Discussions}
\label{subsec_discuss}
    In the subspace learning, the selection of identity representation and the number of subspaces are important.
    In this part, we give some discussions on them.

    For the selection of ${g\left( c \right)}$ in Eqn.\ref{eqn_identity_representation_cls}, we use the average of all image representation in that identity as the identity representation.
    Due to the large variations in viewpoint, illumination and expression, the average operation can remove the individual difference and extract the common characteristics of an identity.
    Actually, ${g\left( c \right)}$ can be considered as a cluster center in the k-means clustering, but with only one center in each identity.

    For the number of subspaces $M$, it cannot be too large or too small, \emph{i.e.,} each subspace should contain a moderate number of identities.
    If $M$ is too small, the whole image space only has a rough division, thus many dissimilar identities will belong to one subspace and not enough hard triplets can be found.
    If $M$ is too large, many small subspaces that contain only a few identities will be generated.
    However, the similarity cannot be guaranteed to be effective as the identity representation is only given by the pretrained classification model, which is not reliable enough to determine the similarity.
    As a result, the fine space division will lead to the over-fitting, which will also give inferior performance.
    In our experiments, we validate that each subspace having about $10k$ identities is appropriate.

\section{Cleaning and Retrieval}
    Except for the triplet network, how to deal with noisy data and large-scale retrieval are also challenging problems.
    In this part, we propose two tricks to handle them, and use the tricks on the MS-Celeb-1M competition.
    Finally, we give the algorithm pipeline as a short summary.

\subsection{Noise Removing}
\begin{figure}[!htb]
\centering
\includegraphics[width=3in]{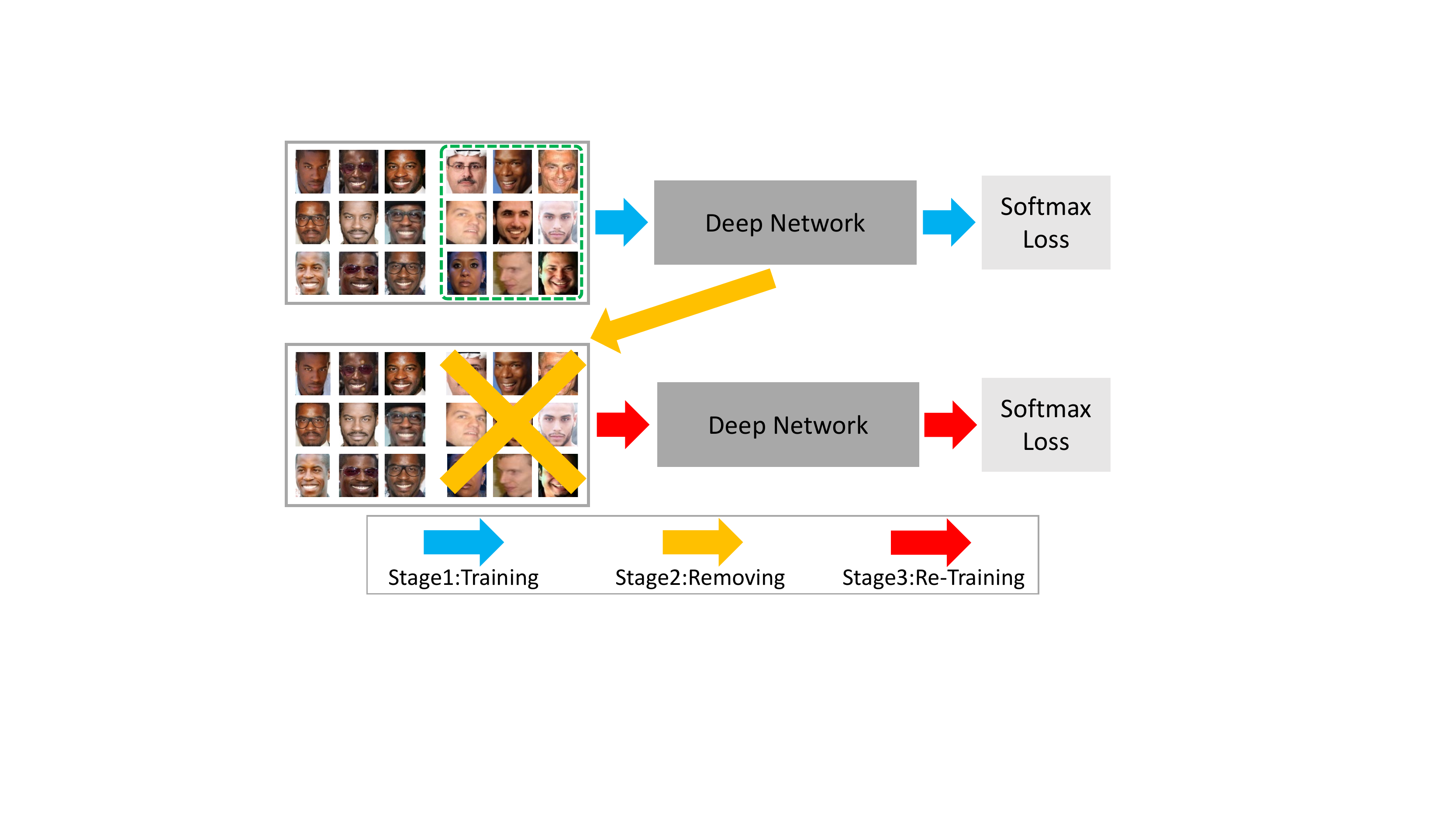}
\caption{An illustration of the noise removing with three steps.}
\label{fg:noise_removal}
\end{figure}

    In $Challenge1$ of the MS-Celeb-1M competition\cite{dataset_msraface}, there are many mislabelled images throughout the dataset.
    We observe that except for some identities that are very noisy, most identities only have a small number of noisy faces.
    As clean data dominates, Sukhbaatar \etal.\cite{cnn_train_noisy} show that CNN can be robust to a small number of noise.
    Based on the evidence, we propose to clean the data with three steps as follows:
    (1) we use all the data to train an initial classification model;
    (2) the initial model is used as feature extractor to remove the noise;
    (3) a new classification model is re-trained with the clean data.
    For the removing step, we adopt a simple trick that only keeps the images with the same predicted identity and labeled identity.
    Fig.\ref{fg:noise_removal} illustrates the removing with three steps.

\subsection{Large-Scale Retrieval}
    As the number of training samples is reduced from $8.4M$ to $5M$ after the cleaning, retrieving a test sample is still challenging.
    Suppose there are $C$ identities with $N$ samples in the training set.
    Directly computing the Euclidean distance to all the training samples gives the complexity of $O\left( N \right)$, but this is infeasible as $N$ is extremely large, and it will take long even with GPU computation.
    To retrieve efficiently, we propose a two-layer hierarchical retrieval with the help of identity representation.
\begin{figure}[!htb]
\centering
\includegraphics[width=3in]{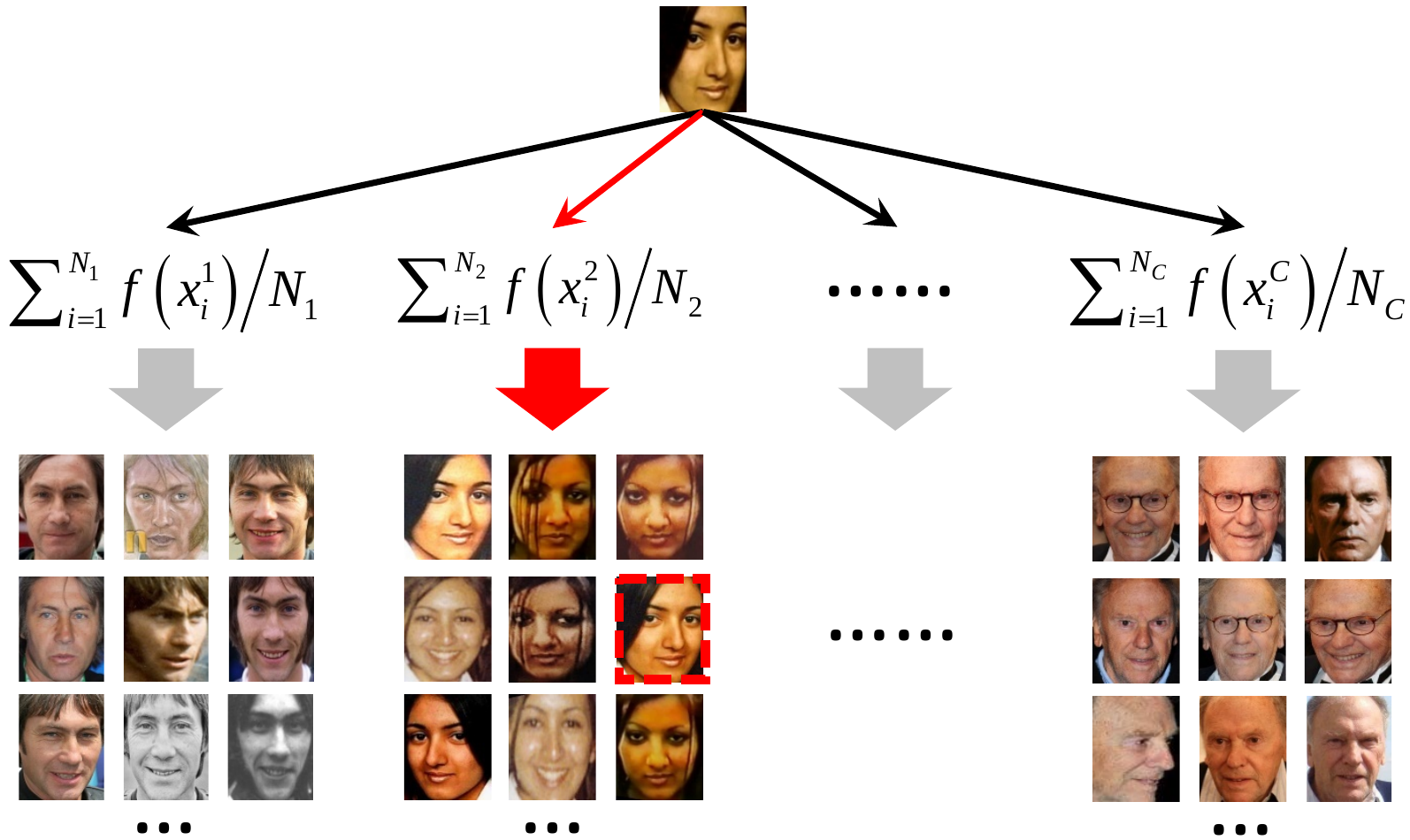}
\caption{An illustration of the proposed two-layer hierarchical retrieval process with identity representation.}
\label{fg:retrieval}
\end{figure}

    Given a test sample, the basic idea is to determine its identity at first, then the training images in that identity are used for retrieval, as shown in Fig.\ref{fg:retrieval}.
    In this way, the complexity can be reduced to only $O\left( {C + {N \mathord{\left/
 {\vphantom {N C}} \right.
 \kern-\nulldelimiterspace} C}} \right)$, wherein ${{N \mathord{\left/
 {\vphantom {N C}} \right.
 \kern-\nulldelimiterspace} C}}$ is the average number of samples in each identity.
    Since there are about $100K$ identities, \emph{i.e.}, $C \ll N$, this hierarchical trick can largely accelerate the retrieval.
    Different from the identity representation in Eqn.\ref{eqn_identity_representation_cls} that uses a pretrained classification model, we adopt the triplet network for representation, which is given by
\begin{equation}
\label{eqn_identity_representation_triplet}
\renewcommand\arraystretch{1}
   {{f\left( c \right) = \sum\nolimits_{i = 1}^{{N_c}} {f\left( {x_i^c} \right)} } \mathord{\left/
   {\vphantom {{f\left( c \right) = \sum\nolimits_{i = 1}^{{N_c}} {f\left( {x_i^c} \right)} } {{N_c}}}} \right.
   \kern-\nulldelimiterspace} {{N_c}}},\;\forall c = 1,...,C.
\end{equation}

\subsection{Algorithm Pipeline}
    As a short summary, we give the pipeline of the triplet with subspace learning in Alg.\ref{alg:alg_pipeline}.
    The pipeline mainly contains three parts: (1) Initialization, which is also the noise removing process; (2) Training, which trains triplet networks with subspace learning; (3) Testing, which is the two-layer hierarchical retrieval.
\renewcommand{\algorithmicrequire}{\textbf{Initilization:}}
\renewcommand{\algorithmicensure}{\textbf{Training:}}
\begin{algorithm}[htbp]
\footnotesize
\caption{The pipeline of triplet with subspace learning.}
\label{alg:alg_pipeline}
\begin{algorithmic}[1]
\REQUIRE ~~\\
    \STATE {Set the margin $\alpha$ and the number of subspaces $M$};
    \STATE {Train an initial classification model on all data or subset};
    \STATE {Remove noisy images by the initial model};
    \STATE {Re-train a classification network on clean data};
\ENSURE ~~\\
    \STATE {Extract image representation $g\left( {x_i^c} \right)$ by the new classification model};
    \STATE {Generate identity representation ${g\left( c \right)}$ for all identities};
    \STATE {Generate $M$ subspaces on all identities with k-means clustering};
    \STATE {Train triplet networks with subspace learning and batch OHNM};
\renewcommand{\algorithmicensure}{\textbf{Testing:}}
\ENSURE ~~\\
    \STATE {Extract image representation $f\left( {x_i^c} \right)$ for all images};
    \STATE {Generate identity representation ${f\left( c \right)}$ for all identities};
    \STATE {Given a test sample, use the two-layer hierarchical retrieval};
\end{algorithmic}
\end{algorithm}

\section{Experimental Evaluation}
    In this section, we give the evaluation of the proposed method.
    We first introduce the experimental setup in detail, then show the main results of training triplet networks, data cleaning and large-scale retrieval.

\subsection{Experimental Settings}
    \emph{\textbf{Database}}: We use two datasets in experiments, including MS-Celeb-1M\cite{dataset_msraface} and LFW\cite{dataset_lfw}.
    The MS-Celeb-1M consists of three parts: training set, development set and test set.
    Firstly, in the training set, there are $99891$ identities with about $8.4M$ images, while data cleaning reduces the number to $5M$.
    Then, in the development set, there are two subsets based on different recognition difficulty: random set(easy samples) and hard set(hard samples), each of which has $500$ images for model selection.
    Finally, in the test set, there are $50K$ images with only $75\%$ identities contained in the training set, and it is used for final evaluation.
    For fair comparison with other methods, we also give results on LFW, wherein the test set has $6000$ image pairs with each pair having the same identity or not.

    \emph{\textbf{Evaluation}}:
    For the evaluation of MS-Celeb-1M\cite{dataset_msraface}, assume there are $N$ images in the development or test set.
    If an algorithm recognizes $M$ images among which $C$ images are correct, the precision and coverage will be calculated as $Precision = {C \mathord{\left/
 {\vphantom {C M}} \right.
 \kern-\nulldelimiterspace} M}$ and $Coverage = {M \mathord{\left/
 {\vphantom {M N}} \right.
 \kern-\nulldelimiterspace} N}$ respectively.
    By varying the confidence threshold, the coverage when $precision=0.95\;or\;0.99$ can be determined.
    For the evaluation of LFW\cite{dataset_lfw}, the accuracy is given by how many pairs are correct in the $6000$ pairs.

    \emph{\textbf{Pre-processing}}:
    In training, we use the same pre-processing in all the networks.
    Given an training image, we first resize it to $256\times256$, then a sub-image with $224\times224$ is randomly cropped and flipped.
    Particularly, we use no mean subtraction or image whitening, as we put a batch normalization layer right after the input data to learn the normalization parameters.
    In the testing phase, both training and testing images are resized to $224\times224$ and flipped, then the average of the original and flipped image representation is considered as the final representation.

    \emph{\textbf{Network and Training}}:
    To learn the large number of identities, we use the popular ResNet-50\cite{cnn_resnet} network, which is deep enough to handle our problem.
    We use a single machine with 4 Titan X in training, and the batchsize is set to be $336$ and $160$ in the classification and triplet network respectively.
    Particularly, the Nesterov Accelerated Gradient(NAG) is adopted for optimization, which is found to converge much quickly than SGD.

    \emph{\textbf{Parameter Setup}}:
    For the learning rate, $0.1$ is used for the classification network that is trained from scratch; while for the triplet network, $0.01$ is used to fine-tune the classification model.
    The training of both networks ends with the rate of $0.0001$, and $20$ epochs are used in each rate to fully converge.
    Then, for the number of subspaces($M$ in Eqn.\ref{eqn_triple_batch_ohem_subspace}), we set $M=10$ to include about $10K$ identities in each subspace.
    Finally, we set the margin $\alpha$ by cross-validation, \emph{i.e.}, $\alpha=0.4$ throughout the experiments.

\subsection{Results of Triplet}
    In this part, we give an evaluation of the triplet network with subspace learning and batch OHNM.
    Table.\ref{table_softmax_triplet_lfw} shows the LFW accuracy of the networks trained with softmax loss and triplet loss.
    Particularly, all the triplet networks are initialized with the pretrained classification model, and the ones with ``\emph{+Softmax}" are fine-tuned jointly with softmax loss and triplet loss.
\vspace{-0.2cm}
\begin{table}[htbp]
\scriptsize
\renewcommand{\arraystretch}{1.1}
\arrayrulewidth=0.5pt \tabcolsep=5pt
\centering
\label{table_softmax_triplet_lfw}
\begin{tabular}{ll}
\toprule
\toprule
Method                              & LFW Acc(\%) \\
\toprule
\toprule
Softmax (Baseline, $100K$)                  & 99.36       \\
\midrule
\midrule
Triplet+Batch OHNM ($100K$)                  & 98.98       \\
Triplet+Batch OHNM+Random-Subspace ($100K$)         & 99.08       \\
\midrule
Triplet+Batch OHNM+Subspace-5 ($100K$)         & 99.25       \\
\textbf{Triplet+Batch OHNM+Subspace-10 ($100K$)}         & \textbf{99.38}       \\
Triplet+Batch OHNM+Subspace-20 ($100K$)         & 99.33       \\
\midrule
\midrule
Triplet+Batch OHNM + Softmax ($100K$)       & 99.33        \\
Triplet+Batch OHNM+Random-Subspace + Softmax ($100K$) & 99.35     \\
\midrule
Triplet+Batch OHNM+Subspace-5 + Softmax ($100K$) & 99.38     \\
\textbf{Triplet+Batch OHNM+Subspace-10 + Softmax} ($100K$) & \textbf{99.48}     \\
Triplet+Batch OHNM+Subspace-20 + Softmax ($100K$) & 99.41     \\
\midrule
FaceNet\cite{cnn_triplet_facenet} (Triplet+OHNM, $8M$)   & 99.63       \\
\bottomrule
\bottomrule
\end{tabular}
\caption{The LFW accuracy of the baseline classification network and different triplet networks.}
\vspace{-0.1cm}
\end{table}

    We have five main observations.
    Firstly, compared to the classification model, the performance of the triplet network with OHNM has decreased a lot, \emph{e.g.}, from $99.36\%$ to $98.98\%$.
    Due to our single machine can only hold small batchsize, the chance of sampling similar identities from $100K$ identities with batch OHNM is very tiny.
    As a result, the network cannot generate enough hard triplets, and it will be trained to fit the semi-hard triplets, which will result in the drop of performance.

    Secondly, the performance of the triplet network with subspace learning increases a little, \emph{e.g.,} from $99.36\%$ to $99.38\%$.
    With enough hard triplets generated, the model is able to overcome the ambiguity in recognizing similar identities.
    However, the improvement is not obvious as we observe the softmax loss increases a lot in training, which indicates that training with only triplet loss will harm the identity information.

    Thirdly, we also compare the triplet network with random subspace, named as \emph{Triplet+Batch OHNM+Random-Subspace}, which divides the whole image space randomly.
    Table.\ref{table_softmax_triplet_lfw} shows that \emph{Random-Subspace} has obtained a slight improvement over the single \emph{Batch OHNM}, \emph{e.g.}, from $98.98\%$ to $99.08\%$,  but it is still far below the subspace learning.
    This result implies that some similar or semi-similar identities may be luckily put together, and the searching space can be reduced a little.
    However, compared to the cluster based subspace learning, not enough hard triplets can be generated.
\begin{figure}[!htb]
\centering
\includegraphics[width=3in]{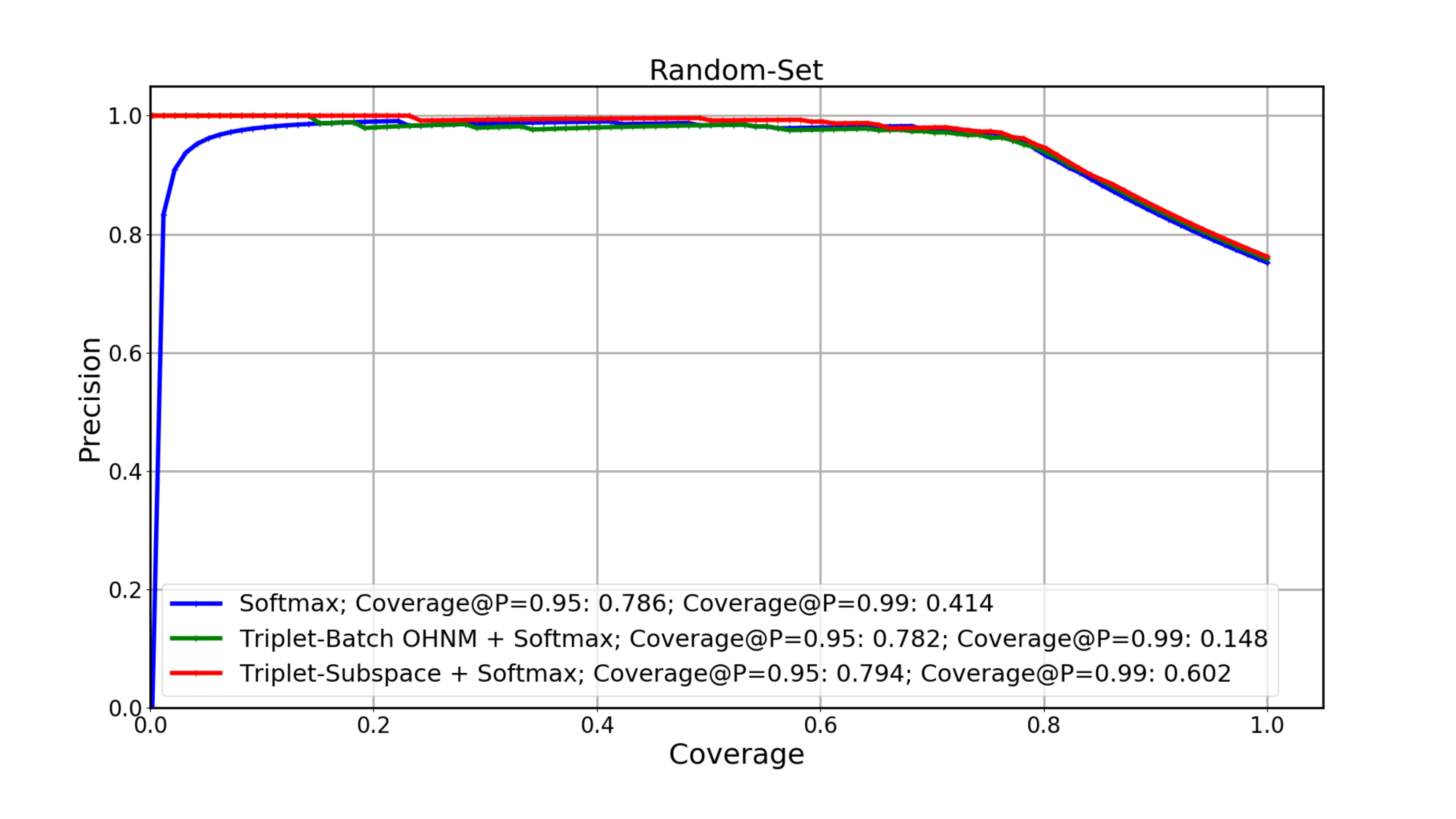}
\caption{The performance of single classification or triplet models on the random set, which belongs to the development set of $Challenge1$ without external data. Best viewed in color.}
\label{fg:random_cov_softmax_triplet}
\vspace{-0.3cm}
\end{figure}
\begin{figure}[!htb]
\centering
\includegraphics[width=3in]{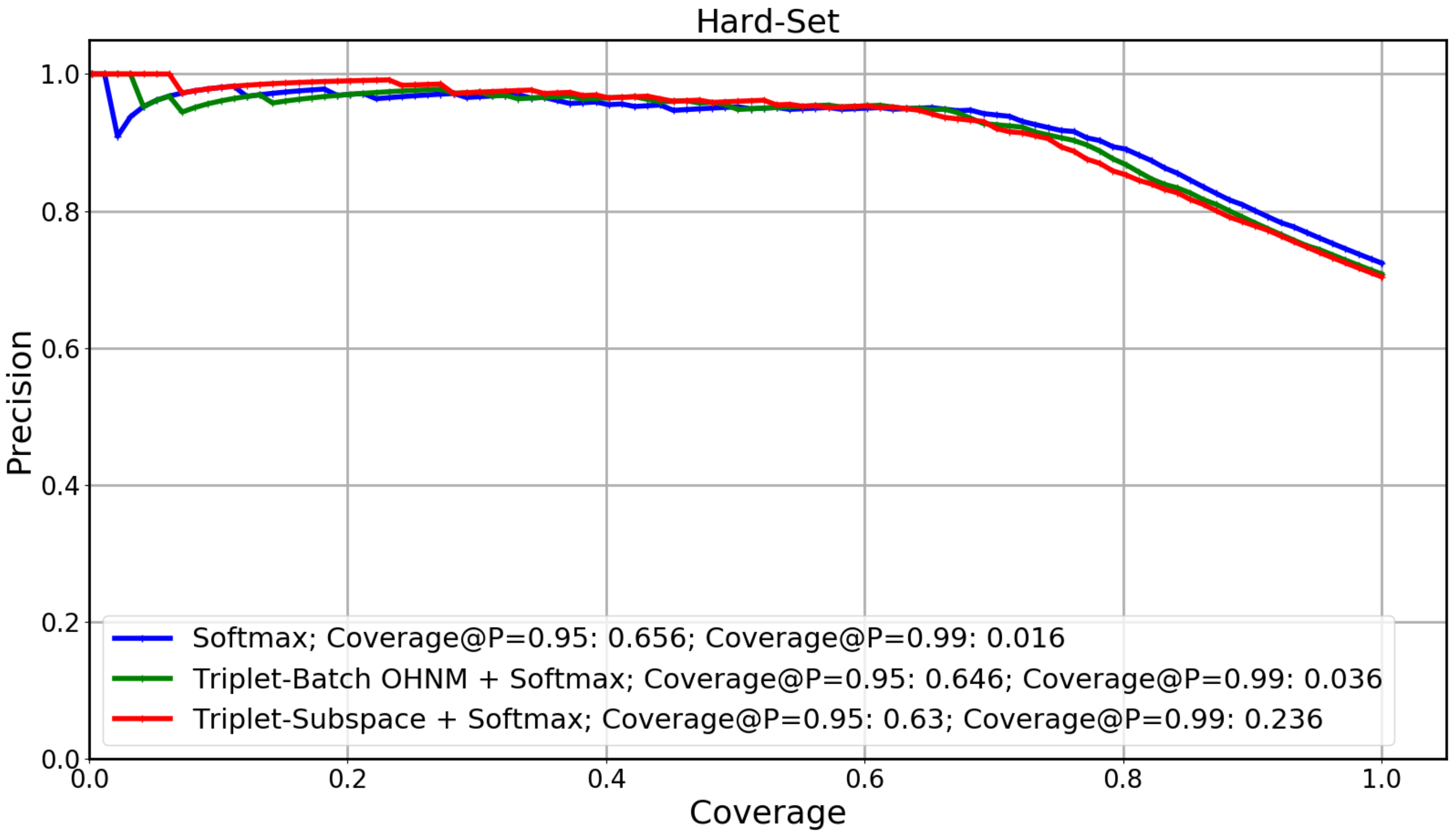}
\caption{The performance of single classification or triplet models on the hard set, which belongs to the development set of $Challenge1$ without external data. Best viewed in color.}
\label{fg:hard_cov_softmax_triplet}
\vspace{-0.5cm}
\end{figure}

    Fourthly, we evaluate the influence of the number of subspaces $M$.
    Three values are tested: $5$, $10$ and $20$, and their corresponding models are denoted as $Subspace-5/10/20$ respectively.
    Table.\ref{table_softmax_triplet_lfw} shows that setting $M=10$ achieves the best performance, \emph{i.e.,} $99.38\%$ for \emph{Triplet+Batch OHNM+Subspace-10}, and this result is just as expected.
    As analyzed in Sec.\ref{subsec_discuss}, small $M$ can only give rough subspaces, each of which contains many dissimilar identities; large $M$ will give too fine division, but this division is only given by the baseline \emph{Softmax} that is not reliable enough, thus over-fitting may happen.

    Finally, compared to the networks trained with only triplet loss, the one trained with additional softmax loss obtains promising improvements, \emph{e.g.}, $99.38\%$ to $99.48\%$ in subspace learning, which can be competitive to FaceNet's $99.63\%$\cite{cnn_triplet_facenet} with $8M$ identities.
    The joint training with softmax and triplet loss is more advantageous because softmax loss can preserve inter-class information while triplet loss can preserve intra-class information, thus the jointly trained model can be more robust.

    Fig.\ref{fg:random_cov_softmax_triplet} and Fig.\ref{fg:hard_cov_softmax_triplet} show the performance of classification and triplet networks on the random and hard set respectively, and the performance is given by the coverage under the precision of $0.95$ and $0.99$.
    It can be observed that \emph{Triplet-Subspace+Softmax} obtains a large improvement over the one with only batch OHNM, especially the coverage under the precision of $0.99$, \emph{e.g.}, from $0.148$ to $0.602$ in the random set and from $0.036$ to $0.236$ in the hard set.
    Besides, the performance in the random set is much higher than the one in the hard set, and this is reasonable as the samples in the hard set are more difficult.

\subsection{Results of Multi-Model}
    To give the best performance, we use two models for evaluation: the baseline classification model(\emph{Softmax} in Table.\ref{table_softmax_triplet_lfw}), and the jointly trained triplet model (\emph{Triplet+Softmax} in Table.\ref{table_softmax_triplet_lfw}).
    We adopt a simple combination that averages the image representation of both models, thus the final representation for the image and identity is given as
\begin{equation}
\label{eqn_multi_model}
\renewcommand\arraystretch{1}
   \frac{{g\left( {x_i^c} \right) + f\left( {x_i^c} \right)}}{2},\quad \frac{{g\left( c \right) + f\left( c \right)}}{2}.
\end{equation}

\begin{figure}[!htb]
\centering
\includegraphics[width=3in]{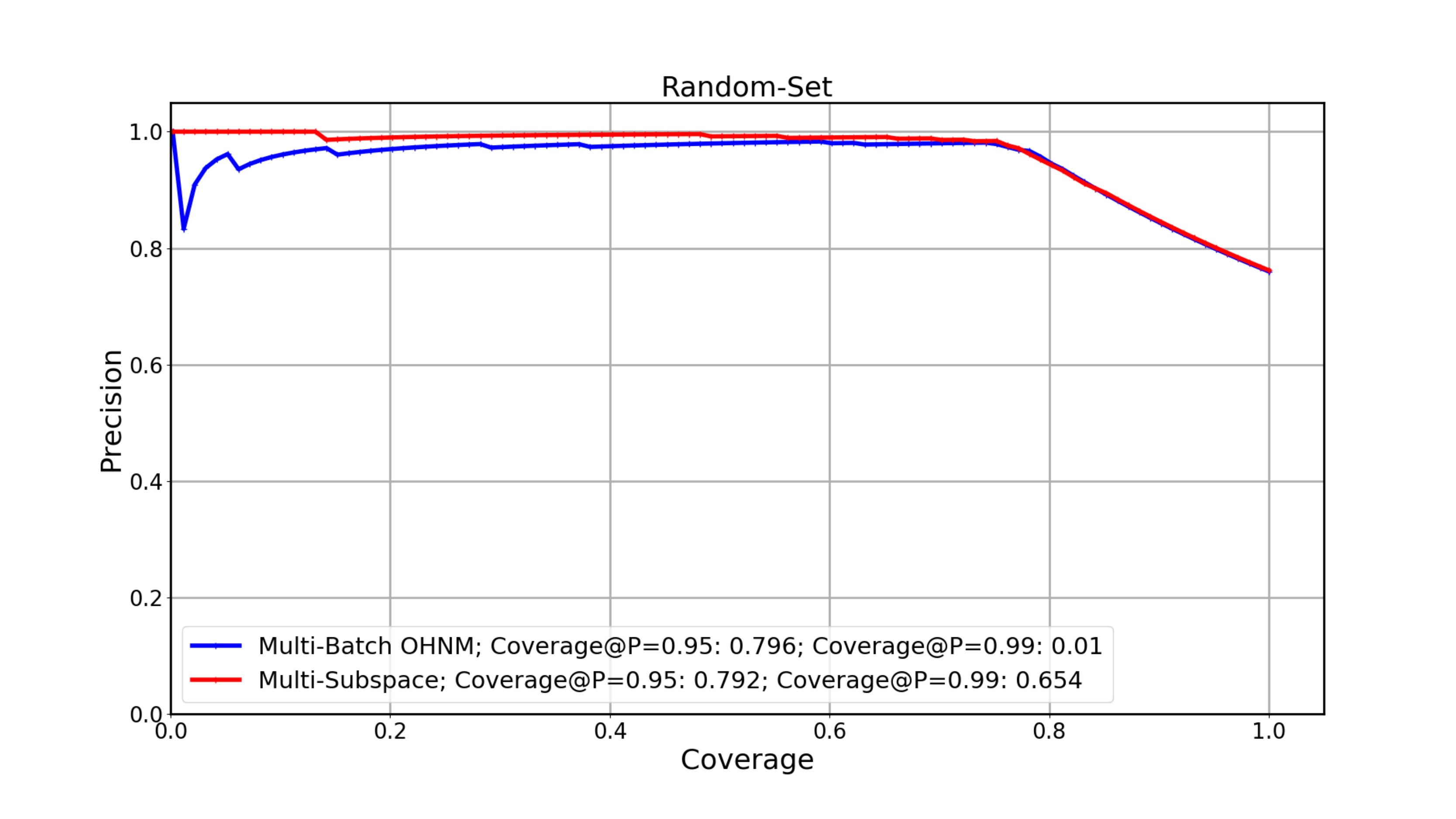}
\caption{The performance of multi-models on the random set, which belongs to the development set of $Challenge1$ without external data. Best viewed in color.}
\label{fg:random_cov_multi}
\vspace{-0.3cm}
\end{figure}

\begin{figure}[!htb]
\centering
\includegraphics[width=3in]{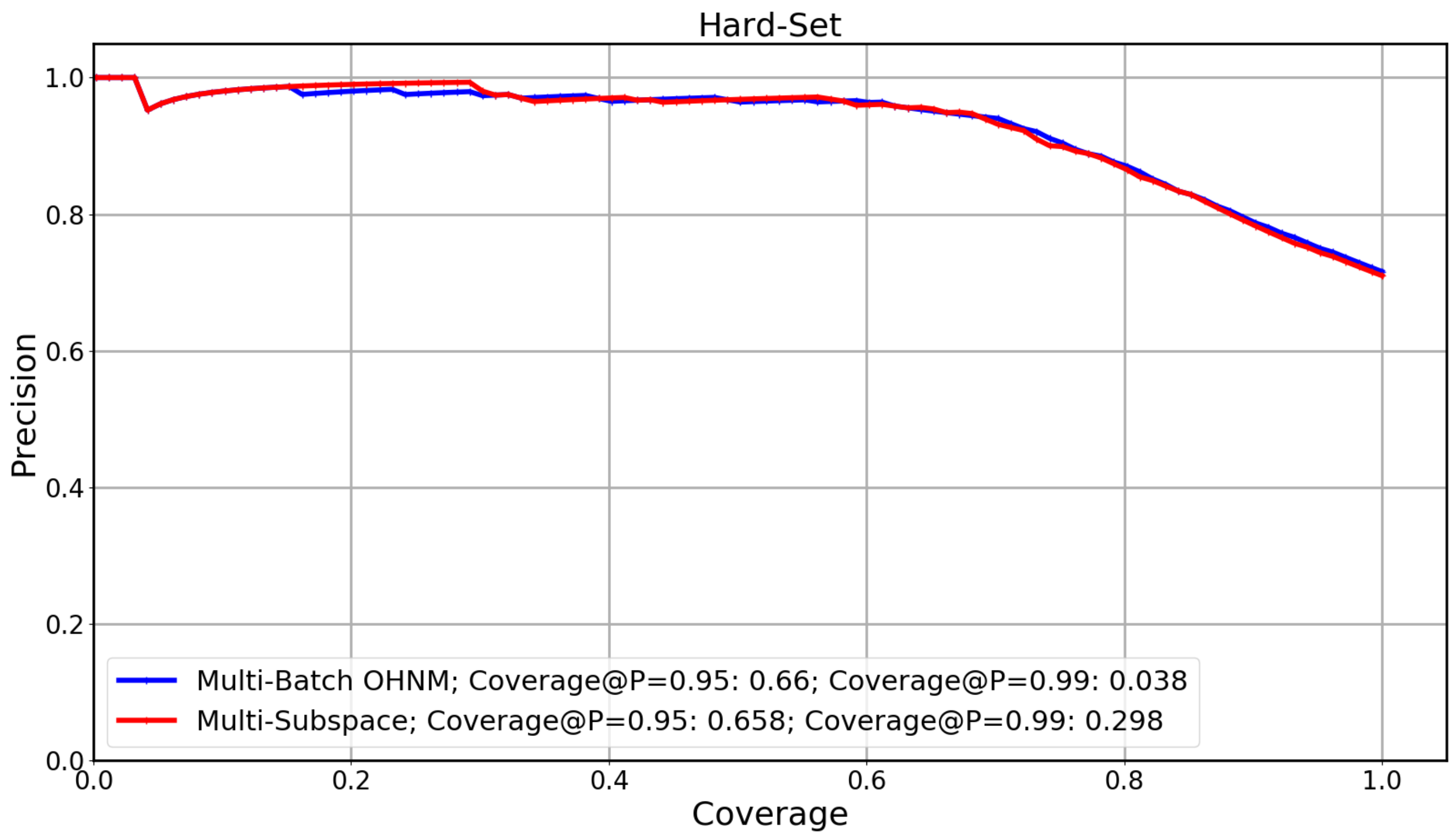}
\caption{The performance of multi-models on the hard set, which belongs to the development set of $Challenge1$ without external data. Best viewed in color.}
\label{fg:hard_cov_multi}
\vspace{-0.1cm}
\end{figure}

    Fig.\ref{fg:random_cov_multi} and Fig.\ref{fg:hard_cov_multi} show the performance of multi-models on the random and hard set respectively, and the performance is given by the coverage under the precision of $0.95$ and $0.99$.
    Particularly, \emph{Multi-Batch OHNM} is the combination of \emph{Softmax} and \emph{Triplet+Batch OHNM+Softmax}, while \emph{Multi-Subspace} consists of \emph{Softmax} and \emph{Triplet+Batch OHNM+Subspace-10+Softmax}.

    It can be observed that \emph{Multi-Subspace} has improved a lot over the single models in Fig.\ref{fg:random_cov_softmax_triplet}, \emph{e.g.}, the coverage increases from $0.602$ to $0.654$ in the random set and from $0.236$ to $0.298$ in the hard set under the precision of $0.99$.
    This demonstrates that the inter-class and intra-class information learned in the classification and triplet network respectively can be complementary to enhance the recognition ability.
    However, we do not see the same improvement for \emph{Multi-Batch OHNM}.

    Compared to the single models in Fig.\ref{fg:random_cov_softmax_triplet}, the performance of \emph{Multi-Batch OHNM} has decreased a lot, \emph{e.g.}, the coverage decreases from the baseline $0.414$ to $0.01$ under the precision of $0.99$ in the random set, while the coverage in the hard set remains basically the same.
    In Fig.\ref{fg:random_cov_softmax_triplet}, the coverage of \emph{Softmax} has dropped a lot as the confidence score increases, which implies that the classification model is not confident to differentiate similar identities, while it is more confident to recognize semi-similar identities as precision is high when the coverage ranges from $0.2\sim0.6$.
    This comes from the fact that \emph{Softmax} focuses more on the inter-class information, but misses details in similar identities.
    Compared to \emph{Softmax}, \emph{Triple-Batch OHNM+Softmax} performs much more confident as high precision can be achieved under a high confidence score, but the precision is lower than \emph{Softmax} when the converge ranges from $0.2\sim0.6$, and this may be caused by the fact that the triplet network with batch OHNM focuses more on the hard triplets.
    As a result, the average of the two models increases the precision in low coverage and decreases the precision in mid-level coverage, which results in the large drop of performance under the precision of $0.99$.
\begin{table}[htbp]
\scriptsize
\renewcommand{\arraystretch}{1.1}
\arrayrulewidth=0.5pt \tabcolsep=1pt
\centering
\label{table_multi_final_evaluation}
\begin{tabular}{lll|lll}
\toprule
\toprule
TeamName      & Data    & Cov@P=0.95 & TeamName      & Data    & Cov@P=0.95 \\
\toprule
\toprule
\textbf{Orion}         & Aligned & \textbf{0.75}          & \textbf{Orion}         & Aligned & \textbf{0.606}         \\\midrule
DRNfLSR       & Aligned & 0.734         & CIGIT\_NLPR   & Aligned & 0.534         \\
ITRC-SARI     & Aligned & 0.707         & DRNfLSR       & Aligned & 0.486         \\
CIGIT\_NLPR   & Aligned & 0.684         & faceman       & Aligned & 0.33          \\
ms3rz         & Aligned & 0.646         & ms3rz         & Aligned & 0.26          \\
1510          & Aligned & 0.57          & FaceAll       & Aligned & 0.254         \\
FaceAll       & Aligned & 0.554         & BUPT\_PRIS    & Aligned & 0.21          \\
faceman       & Aligned & 0.461         & IMMRSB3RZ     & Aligned & 0.042         \\
BUPT\_PRIS    & Aligned & 0.421         & BUPT\_MCPRL        & Cropped & 0.04          \\
IMMRSB3RZ     & Aligned & 0.171         & CIIDIP     & Aligned & 0.02          \\
CIIDIP        & Aligned & 0.154         & ITRC-SARI       & Aligned & 0.004         \\
BUPT\_MCPRL   & Cropped & 0.064         & DS\_NFS          & Aligned & 0.001         \\
NII-UIT-KAORI & Aligned & 0.001         & 1510     & Aligned & 0.001         \\
DS\_NFS       & Aligned & -             & Paparazzi & Aligned & -             \\
Paparazzi     & Aligned & -             & NII-UIT-KAORI & Aligned & -             \\
\bottomrule
\bottomrule
\end{tabular}
\caption{The final results on the random set(left) and hard set(right) in $Challenge1$ without using external data.
The performance is given by the coverage under the precision of $0.95$.}
\end{table}

    For the final evaluation, we adopt the \emph{Multi-Subspace} for result submission.
    Table.\ref{table_multi_final_evaluation} shows the final results of our method(Orion) and other teams on the random set(left) and hard set(right) in $Challenge1$ without using external data, and the performance is given by the coverage under the precision of $0.95$.
    It can be clearly observed that our method has achieved the state-of-the-art performance in both the random set and hard set, and the coverage in the hard set has improved a lot over other teams, \emph{e.g.}, from $0.534$ by \emph{CIGIT\_NLPR} to our $0.606$.
    Particularly, in the random set, we have obtained the coverage of $0.75$, which is the upper limit in $Challenge1$ without using external data as only $75\%$ training identities are included in the test set.
    This result is surprising because we have obtained $100\%$ recall on the training identities under the precision of $0.95$.
    Table.\ref{table_multi_final_evaluation_external} also shows the final results on the random and hard sets using the external data.
    Even though no external data is used in our method, we can achieve competitive performance to some teams, \emph{e.g.}, the \emph{SmileLab}.
\vspace{-0.2cm}
\begin{table}[htbp]
\scriptsize
\renewcommand{\arraystretch}{1}
\arrayrulewidth=0.5pt \tabcolsep=2pt
\centering
\label{table_multi_final_evaluation_external}
\begin{tabular}{l|l|l|l|l}
\toprule
\toprule
External & Team Name     & Data    & Cov@P=0.95\_Random & Cov@P=0.95\_Hard \\
\toprule
\toprule
Yes      & Panasonic-NUS & Aligned & 0.875              & 0.791            \\
Yes      & Turtle        & Aligned & 0.862              & 0.73             \\
Yes      & SmileLab      & Aligned & 0.792              & 0.61             \\
\midrule
\textbf{No}       & Orion         & Aligned & 0.75               & 0.606            \\
\midrule
Yes      & D**           & Cropped & 0.745              & 0.454            \\
Yes      & BMTV          & Cropped & 0.724              & 0.409            \\
Yes      & FaceSecret    & Aligned & 0.641              & 0.002            \\
\bottomrule
\bottomrule
\end{tabular}
\caption{The final results on the random set and hard set in $Challenge1$ with external data.
The performance is given by the coverage under the precision of $0.95$.}
\end{table}

\subsection{Results of Cleaning}
    In this part, we give an evaluation of the noise removing method.
    Table.\ref{table_noise_lfw} shows the LFW accuracy of three different noisy removing methods on MS-Celeb-1M, and a classification model with softmax loss is used to remove noise.
    \emph{Original Data} uses all the data in training; while \emph{Fixed Ratio} uses other clean datasets to train a model as feature extractor, which keeps a fixed ratio of images in each identity, and this is used in last year's competition.
    It can be observed that our removing method obtains the LFW accuracy of $99.36\%$, which is much higher than other methods, and Fig.\ref{fg:noise_removal} shows an example.
    Besides, we see that directly using all the data in training is even better than \emph{Fixed Ratio}, and this indicates CNN can be robust to a small number of noises, as demonstrated in \cite{cnn_train_noisy}.
    Based on our observation, different identities have different number of noises.
    As a result, \emph{Fixed Ratio} will remove many clean samples for clean identities and keep many noisy samples in noisy identities, thus it may fail.
\begin{table}[htbp]
\scriptsize
\renewcommand{\arraystretch}{1.2}
\arrayrulewidth=0.5pt \tabcolsep=13pt
\centering
\label{table_noise_lfw}
\begin{tabular}{l|lll}
\toprule
\toprule
Method & Original Data & Fixed Ratio   & Ours  \\
\midrule
LFW Acc(\%)&  98.96    & 98.85 & 99.36 \\
\bottomrule
\bottomrule
\end{tabular}
\caption{The LFW accuracy based on three different noisy removing methods on the training set of MS-Celeb-1M.}
\end{table}
\begin{figure}[!htb]
\centering
\includegraphics[width=3in]{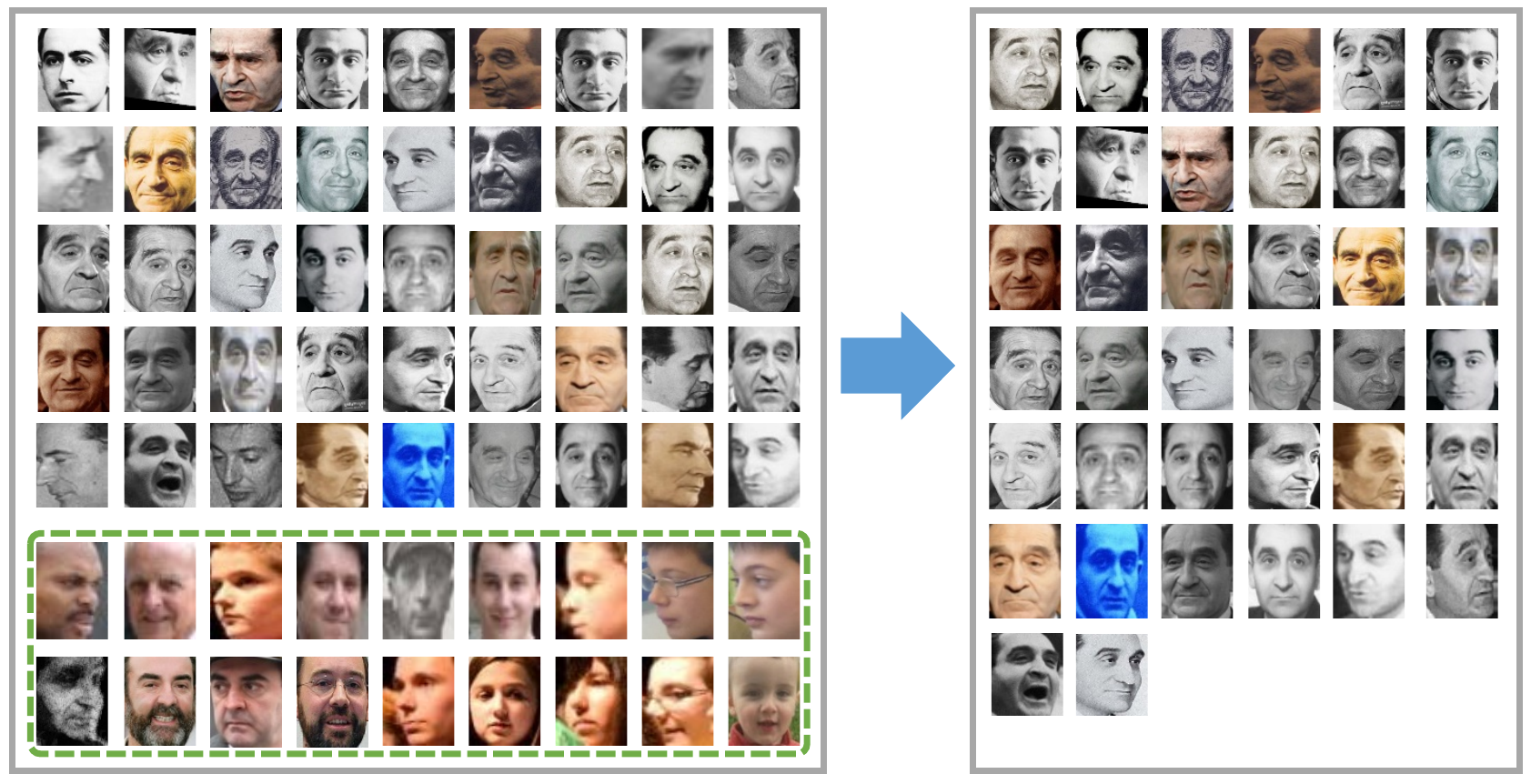}
\caption{Results of noise removing in two identities. The images in the green dotted rectangles are noise faces.}
\label{fg:result_noise_removal}
\vspace{-0.3cm}
\end{figure}

\subsection{Results of Retrieval}
    In this part, we give an evaluation of the retrieval efficiency.
    In testing, we have $99891$ identities with about $5.04M$ images in the retrieval set, and the objective is to determine the identity for a given test image.
    Table.\ref{table_retrieval_time} gives the time cost(s) of retrieving one image with different retrieval methods.
    \emph{All} directly calculates the Euclidean distance to all samples, $2-Hierarchy$ denotes our two-layer hierarchical retrieval trick, and CPU/GPU are the different hardware implementations.
    It can be observed that with the CPU implementation, our trick can largely accelerate the retrieval by $69 \times$.
    This is as expected as $All$ needs $5.04M$ multiplications, while $2-Hierarchy$ only needs $99891 + 50$ multiplications, which can save a lot of time.
    To further accelerate the retrieval, we implement matrix multiplication by GPU and the time is reduced to only $0.052s$ for each image, which is very efficient.
    In the final evaluation, we adopt $2-Hierarchy$ with $GPU$ to finish the testing with $50K$ images in about $1$ hour(including feature extraction and retrieval).
\begin{table}[htbp]
\scriptsize
\renewcommand{\arraystretch}{1}
\arrayrulewidth=0.5pt \tabcolsep=15pt
\centering
\label{table_retrieval_time}
\begin{tabular}{l|ll}
\toprule
\toprule
Method & CPU Time(s) & GPU Time(s)  \\
\midrule
All&  32.09    &  4.646 \\
2-Hierarchy&  0.464    & 0.052 \\
\bottomrule
\bottomrule
\end{tabular}
\caption{The time cost(s) of retrieving one image with different retrieval methods and CPU/GPU implementations.}
\vspace{-0.4cm}
\end{table}

\section{Conclusion}
    In this paper, we take face recognition as a breaking point and study how to train triplet networks with large-scale data.
    Firstly, to find similar identities for more effective hard triplets, we have proposed the triplet with subspace learning, and this is our major contribution.
    Then, to handle noisy faces and large-scale retrieval, we propose two tricks that use a three-step noise removing trick and a two-layer hierarchical retrieval trick.
    Combined with the subspace learning, we have achieved the state-of-the-art performance on the MS-Celeb-1M $Challenge1$ without external data.
    In our future work, we will study triplet in a larger scale, \emph{e.g.}, $1M$ or $10M$ identities.

\section*{Acknowledgement}
    We would like to thank Huajiang Xu and Yangang Zhang for their discussions and supports on data processing and computing architectures.


\begin{thebibliography}{10}\itemsep=-1pt

\bibitem{cnn_triplet_multi-task}
W.~Chen, X.~Chen, J.~Zhang, and K.~Huang.
\newblock A multi-task deep network for person re-identification.
\newblock In {\em AAAI}, 2017.

\bibitem{dataset_msraface}
Y.~Guo, L.~Zhang, Y.~Hu, X.~He, and J.~Gao.
\newblock Ms-celeb-1m: A dataset and benchmark for large-scale face
  recognition.
\newblock In {\em ECCV}, 2016.

\bibitem{cnn_resnet}
K.~He, X.~Zhang, S.~Ren, and J.~Sun.
\newblock Deep residual learning for image recognition.
\newblock In {\em CVPR}, 2016.

\bibitem{cnn_triplet_defense}
A.~Hermans, L.~Beyer, and B.~Leibe.
\newblock In defense of the triplet loss for person re-identification.
\newblock In {\em arXiv:1703.07737}, 2017.

\bibitem{cnn_triplet_to_match_cls}
E.~Hoffer and N.~Ailon.
\newblock Deep metric learning using triplet network.
\newblock In {\em ICLR}, 2015.

\bibitem{dataset_lfw}
E.~Learned-Miller, G.~B. Huang, A.~RoyChowdhury, and G.~Hua.
\newblock Labeled faces in the wild: A survey.
\newblock In {\em Advances in Face Detection and Facial Image Analysis}, 2016.

\bibitem{cnn_triplet_vggface}
O.~M. Parkhi, A.~Vedaldi, and A.~Zisserman.
\newblock Deep face recognition.
\newblock In {\em BMVC}, 2015.

\bibitem{cnn_train_noisy}
M.~P. L. B. R.~F. S.~Sukhbaatar, J.~Bruna.
\newblock Training convolutional networks with noisy labels.
\newblock In {\em ICLR Workshop}, 2015.

\bibitem{cnn_triplet_to_probilistic_embed_cls}
S.~Sankaranarayanan, A.~Alavi, C.~Castillo, and R.~Chellappa.
\newblock Triplet probabilistic embedding for face verification and clustering.
\newblock In {\em ICLR}, 2015.

\bibitem{cnn_triplet_facenet}
F.~Schroff, D.~Kalenichenko, and J.~Philbin.
\newblock Facenet: A unified embedding for face recognition and clustering.
\newblock In {\em CVPR}, 2015.

\bibitem{cnn_vggnet}
K.~Simonyan and A.~Zisserman.
\newblock Very deep convolutional networks for large-scale image recognition.
\newblock In {\em ICLR}, 2015.

\bibitem{cnn_triplet_lift_structure}
H.~O. Song, Y.~Xiang, S.~Jegelka, and S.~Savarese.
\newblock Deep metric learning via lifted structured feature embedding.
\newblock In {\em CVPR}, 2016.

\bibitem{cnn_triplet_early_online}
J.~Wang, Y.~Song, T.~Leung, C.~Rosenberg, J.~Wang, J.~Philbin, B.~Chen, and
  Y.~Wu.
\newblock Learning fine-grained image similarity with deep ranking.
\newblock In {\em CVPR}, 2014.

\bibitem{cnn_deep_retrieval}
F.~Zhao, Y.~Huang, L.~Wang, and T.~Tan.
\newblock Deep semantic ranking based hashing for multi-label image retrieval.
\newblock In {\em CVPR}, 2015.

\bibitem{cnn_triplet_to_binary_cls}
B.~Zhuang, G.~Lin, C.~Shen, and I.~Reid.
\newblock Fast training of triplet-based deep binary embedding networks.
\newblock In {\em CVPR}, 2016.

\end{thebibliography}
\end{document}